# Co-CoT: A Prompt-Based Framework for Collaborative Chain-of-Thought Reasoning

**Seunghyun Yoo**
Carter Student AI Research
seunghyun050616@gmail.com

**Abstract**

Due to the proliferation of short-form content and the rapid adoption of AI, opportunities for deep, reflective thinking have significantly diminished—undermining users' critical thinking and reducing engagement with the reasoning behind AI-generated outputs. To address this issue, we propose an Interactive Chain-of-Thought (CoT) Framework that enhances human-centered explainability and responsible AI usage by making the model's inference process transparent, modular, and user-editable. The framework decomposes reasoning into clearly defined blocks that users can inspect, modify, and re-execute, encouraging active cognitive engagement rather than passive consumption. It further integrates a lightweight edit-adaptation mechanism inspired by preference learning, allowing the system to align with diverse cognitive styles and user intentions. Ethical transparency is ensured through explicit metadata disclosure, built-in bias checkpoint functionality, and privacy-preserving safeguards. This work outlines the design principles and architecture necessary to promote critical engagement, responsible interaction, and inclusive adaptation in AI systems aimed at addressing complex societal challenges.

## 1 Introduction

The growing dominance of short-form digital content and widespread integration of AI technologies have reshaped how individuals consume and interact with information. Platforms such as TikTok, Instagram Reels, and YouTube Shorts now account for significant portions of daily media consumption, with recent studies indicating that U.S. undergraduates spend an average of 95 minutes per day watching short-form video content (Fedak et al., 2024). These highly compressed formats promote the rapid intake of information while minimizing cognitive effort—shifting focus from deep understanding to immediate results. Simultaneously, AI systems, particularly large language models (LLMs), have become widely accessible tools for productivity, learning, and decision-making. However, users frequently interact with these models in a manner that prioritizes outputs over reasoning, further reinforcing a culture of result-oriented engagement. This dual trend—short-form consumption and passive AI use—has contributed to a decline in metacognitive engagement, critical reasoning, and the development of reflective problem-solving skills. In high-stakes domains such as education, healthcare, and governance, the consequences of this shift are especially pronounced: users rely on model outputs without understanding the rationale behind them, leaving room for over-trust, misinterpretation, and overlooked biases.

To mitigate these risks and promote responsible AI integration, we introduce an Interactive Chain-of-Thought (CoT) Framework that empowers users to collaborate with AI models through editable and transparent reasoning steps. Unlike prior prompting techniques such as Chain-of-Thought (Wei et al., 2022) or Tree-of-Thought (Yao et al., 2023), which expose intermediate steps but do not allow user intervention, our approach supports direct modification of the model's reasoning before the final answer is produced. This design allows users to interrogate, revise, and refine the logic of AI systems—facilitating more inclusive, accountable, and human-centered AI applications.

In what follows, we detail the structure of the proposed framework, present its core design principles, and discuss how this approach can foster metacognitive engagement and trustworthy AI integration across domains where ethical reasoning and explainability are essential.

## 2 Related Works

Recent advancements in prompting strategies have significantly enhanced the reasoning capabilities of large language models (LLMs). Notably, Chain-of-Thought (CoT) prompting, introduced by Wei et al. (2022), involves guiding LLMs to generate intermediate reasoning steps, thereby improving performance on complex tasks such as arithmetic and commonsense reasoning.

Building upon CoT, Tree-of-Thought (ToT) prompting, proposed by Yao et al. (2023), enables LLMs to explore multiple reasoning paths in a tree-like structure. This approach allows for deliberate decision-making by considering various possibilities and backtracking when necessary, leading to improved problem-solving abilities in tasks requiring strategic planning.

Further extending the interactivity of reasoning processes, the iToT system developed by Boyle et al. (2024) offers a visual interface that allows users to inspect, correct, and extend the model's thoughts. This interactive system enhances transparency and user engagement in the model's problem-solving process.

In parallel, efforts to align LLM outputs with user preferences have led to the development of frameworks like PRELUDE. This approach learns latent user preferences from direct edits, enabling the model to adapt its responses to better match individual user expectations.

Our proposed Interactive CoT Framework builds upon these advancements by integrating editable reasoning blocks, user-driven adaptations, and ethical transparency measures. This framework aims to foster deeper user engagement, enhance explainability, and promote responsible AI usage in complex decision-making scenarios.

## 3 Methodology

We propose a prompt-driven Interactive Chain-of-Thought (CoT) Framework that enables users to inspect and modify the model's intermediate reasoning steps directly through multi-turn natural language interactions. Rather than relying on a graphical interface, users engage with the model conversationally—inspecting each step, suggesting edits, and requesting regeneration—all within a dialogue loop.

### 3.1 Prompt-Driven CoT Workflow

Given a user query $q$, the language model generates an initial reasoning chain:

[Step 1] …
[Step 2] …
[Step 3] …

The model then prompts:

Model: Do you want to edit any step before I generate the final answer?

Users may respond with:

* Yes, revise Step 2 to consider alternative assumption A′.
* Remove Step 1 and merge Steps 2 and 3.
* Looks good. Proceed to final answer.

The framework re-generates affected steps and continues the reasoning chain. This creates a tight edit–regenerate loop without any additional interface requirements.

### 3.2 Reasoning Block Management via Prompts

To maintain consistency, the model assigns a unique identifier to each step and tracks dependencies. Example interaction:

[Step 2] The agent assumes that X implies Y.
User: Replace Step 2 with: Assume X leads to Z instead.
Model: Updated Step 2 acknowledged. Recalculating Steps 3 and 4...

This prompt-based block-level editing allows users to maintain full control over the evolving reasoning process using natural language alone.

### 3.3 Edit Adaptation Mechanism

We log all (original step, user revision) pairs and apply an online preference-learning component. Future completions are biased toward user-specific revision patterns, leveraging edit histories to adapt model behavior.

This is implemented via:
- Reranking sampled completions to match prior linguistic or structural preferences

- Modifying internal prompting to reflect learned user heuristics (e.g., "User prefers counterexamples over assumptions")

### 3.4 Ethical Safeguards in Dialogue

Our framework includes lightweight safeguards within the conversational flow:

- **Transparency Prompts**: Model explicitly discloses version, parameters, and uncertainty.
- **Bias Challenge Prompt**: At any point, the user may ask, "Is there any bias in Step X?" prompting the model to self-audit or reframe the logic.
- **Privacy Reminders**: System warns if identifiable information is detected in user edits or model outputs.

## 4 Use Case: Co-CoT

The following case study demonstrates how our framework supports ethical reasoning in practice, particularly around fairness and inclusion in AI language systems. We begin with a user question:

> [User]: How can we ensure that language models treat minority dialects fairly?
>
> [Model]:
>
> **[Step 1] Define fairness** in the context of language models and minority dialects.
>
> **[Step 3] Evaluate current model performance** on these dialects.
>
> **[Step 4] Analyze sources of disparity**, such as underrepresentation in training data, biased annotation practices, or homogenizing tokenization strategies.
>
> **[Step 5] Augment and diversify training data** to include more examples of minority dialects in authentic contexts.
>
> **[Step 6] Implement fairness-aware training techniques**, such as adversarial debiasing, data reweighting, or domain adaptation for dialectal variation.
>
> **[Step 7] Use dialect-specific evaluation metrics** to monitor improvements in fairness.
> This includes both quantitative (e.g., perplexity, accuracy) and qualitative (e.g., fluency, respectfulness) measures.
>
> **[Step 8] Engage with affected communities** to co-design evaluation criteria and acceptable use cases.
> **[Step 9] Promote transparency** by documenting dialect coverage and known biases in model cards and usage guidelines.

> [User]: Let's revise Step 4. Instead of just analyzing disparities, let's also recognize that not all dialectal variation stems from underrepresentation.
>
> [Model]:
>
> These updates reflect a more critical lens on how data is collected and labeled, and how remediation must address both absence and distortion.
>
> **[Step 5] Augment and diversify training data** by not only increasing dialectal representation but also correcting past misannotations and filtering out biased moderation decisions. Crowdsourcing and linguistic validation can help ensure dialects are captured respectfully and accurately.
>
> **[Step 6] Implement fairness-aware training techniques** that also account for de-stigmatization—e.g., retraining models to recognize dialectal grammar as valid, and designing classifiers that don't misflag dialects as toxic or low-quality.
>
> **[Step 7] Use dialect-specific evaluation metrics** that capture both performance and bias reduction. Metrics should test whether dialectal inputs are understood without distortion and whether outputs reflect appropriate tone, fluency, and respect.
>
> **[Step 8] Engage with affected communities** not just for validation, but to understand lived experiences of stigmatization and integrate these perspectives into model development and evaluation pipelines.
>
> **[Step 9] Promote transparency** by documenting both the presence and treatment of dialects—e.g., how annotation policies were revised, or what moderation filters were adjusted to avoid dialectal discrimination.

> [User]: I'd like to expand Step 7. Could we include participatory evaluation methods, like having native speakers of the dialect assess model outputs directly?
>
> [Model]:
>
> **[Step 7] Use dialect-specific and participatory evaluation methods** to assess model fairness.
> In addition to quantitative metrics (e.g., accuracy, perplexity, toxicity scores), incorporate evaluations from native speakers of the dialect who can assess fluency, respectfulness, and cultural appropriateness.
> Be cautious with automated metrics, which may misinterpret dialectal grammar or semantics as errors or toxicity due to built-in biases in evaluation tools.

This case study illustrates how Co-CoT enables users to collaboratively construct, revise, and validate complex reasoning processes involving fairness and bias. By transparently walking through each step—and allowing edits before reaching a

conclusion—our framework empowers users to challenge assumptions, correct overlooked nuances (such as dialectal stigmatization), and arrive at more ethically robust outcomes.

Beyond this example, the framework is broadly applicable to a range of ethical concerns in AI, including representation, accountability, and harm mitigation. By combining structured reasoning with user agency, Co-CoT supports not only *better answers*, but also *better questions*.

## 5 Discussion

The Co-CoT framework is designed not only to support ethical oversight, but to actively enhance users' reasoning processes through structured interaction. Rather than providing ready-made answers, it encourages users to critically engage with the assumptions and logical structure behind each conclusion.

By explicitly breaking down reasoning into editable, inspectable steps, Co-CoT acts as a kind of cognitive scaffold—helping users articulate complex ethical concerns, question default framings, and explore alternative perspectives. This is especially valuable in contexts like dialectal fairness, where intuitive notions of correctness may obscure deeper issues of representation, power, or stigma.

Crucially, Co-CoT treats ethical reasoning as an iterative and co-creative process. Users are not passive recipients of "ethical outputs" but active participants in shaping how fairness, bias, and inclusion are understood and applied. In doing so, the system fosters reflective thinking, inviting users to slow down and refine their own mental models alongside the model's outputs.

This approach has broader implications for how we design AI systems—not just as tools for automation, but as partners in human thought. In that sense, Co-CoT points toward a future where AI helps not just to solve problems, but to think better about them.

## 6 Limitation

While Co-CoT provides a flexible and interactive framework for ethical reasoning, several limitations warrant consideration.

First, the framework assumes a certain level of user engagement and cognitive effort. It depends on users being willing and able to critically review and revise reasoning steps. In time-sensitive or high-volume settings, this kind of interaction may feel too slow or demanding, which can limit adoption in fast-paced AI development environments.

Second, although Co-CoT can surface ethical tensions, it does not guarantee resolution. In cases involving deeply conflicting values or entrenched institutional biases, the system can illuminate the problem but may not provide a clear or actionable solution without additional human judgment or support structures.

Third, the effectiveness of the framework depends on the quality of the initial reasoning steps and prompts. If those starting points are biased or poorly constructed, they can shape the entire trajectory of the reasoning, even if revision is allowed. While user edits help mitigate this, not all users may feel confident or equipped to challenge problematic assumptions.

Finally, there are limitations related to scalability and automation. Because the framework emphasizes transparency and deliberation, it is not well suited to fully automated systems or large-scale deployments where human-in-the-loop feedback is impractical.

Despite these challenges, Co-CoT remains a compelling alternative to static or opaque ethical tools, especially in contexts where depth, reflection, and collaboration are more valuable than speed or scale.

# A  Appendices

## A1. Full Prompt Template for Co-CoT Framework

> You are an Interactive Reasoning Assistant designed to guide users through a transparent, multi-step reasoning process.
>
> Your goal is to expose your chain-of-thought reasoning in numbered steps and allow the user to review and modify any step before you generate a final answer.
>
> You must follow this exact workflow:
>
> 1. Generate a clear, step-by-step reasoning chain (label each with [Step 1], [Step 2], etc.).
>
> 2. Present the full reasoning chain to the user and ask if they would like to edit, delete, or replace any step.
>
> 3. If the user provides an edit:
>
>    - Acknowledge the edit.
>
>    - Update the modified step.
>
>    - Automatically re-calculate any logically dependent downstream steps.
>
>    - Show the updated reasoning chain.
>
>    - Ask if further changes are needed.
>
> 4. If the user confirms the chain (e.g., says "Continue"), proceed to generate the final answer based strictly on the most recent version of the reasoning chain.

> Important constraints:
>
> - Never finalize an answer until the user explicitly confirms the reasoning chain.
>
> - Always number steps and allow edits by number (e.g., "Replace Step 3 with…").
>
> - When updating the chain, only modify what logically depends on the user's change.
>
> - Encourage critical review with language such as: "Would you like to revise or reframe any of these steps?"
>
> Do not assume the user wants to continue unless they clearly state so.
>
> At the end, optionally offer to export the reasoning chain and final answer.

5